\documentclass[runningheads]{llncs}
\usepackage[T1]{fontenc}
\usepackage{graphicx}
\usepackage{amsmath,amsfonts,mathtools}
\usepackage{algorithm}
\usepackage{algpseudocode}
\usepackage{float}%
\usepackage{amssymb}
\usepackage{booktabs}
\usepackage{xcolor}
\usepackage[hidelinks]{hyperref} %
\begin{document}
\title{DiabetesNet: A Deep Learning Approach to Diabetes Diagnosis}
\author{Zeyu Zhang\inst{1}\orcidID{0009-0006-8819-3741} \and
Khandaker Asif Ahmed\inst{2}\orcidID{0000-0003-1271-7781} \and
Md Rakibul Hasan\inst{3}\orcidID{0000-0003-2565-5321} \and
Tom Gedeon\inst{1,3, 4}\orcidID{0000-0001-8356-4909} \and
Md Zakir Hossain\inst{1,2,3}\orcidID{0000-0003-1892-831X}}
\authorrunning{Zhang et al.}
\institute{Australian National University, Canberra, Australia \and Commonwealth Scientific and Industrial Research Organisation, Australia \and Curtin University, Perth, Australia \and Obuda University, Hungary}
\maketitle              %
\begin{abstract}
Diabetes, resulting from inadequate insulin production or utilization, causes extensive harm to the body. Existing diagnostic methods are often invasive and come with drawbacks, such as cost constraints. Although there are machine learning models like Classwise k Nearest Neighbor (CkNN) and General Regression Neural Network (GRNN), they struggle with imbalanced data and result in under-performance. Leveraging advancements in sensor technology and machine learning, we propose a non-invasive diabetes diagnosis using a Back Propagation Neural Network (BPNN) with batch normalization, incorporating data re-sampling and normalization for class balancing. Our method addresses existing challenges such as limited performance associated with traditional machine learning. Experimental results on three datasets show significant improvements in overall accuracy, sensitivity, and specificity compared to traditional methods. Notably, we achieve accuracies of \textbf{89.81\%} in Pima diabetes dataset, \textbf{75.49\%} in CDC BRFSS2015 dataset, and \textbf{95.28\%} in Mesra Diabetes dataset. This underscores the potential of deep learning models for robust diabetes diagnosis. See project website
\textbf{\textcolor{blue}{\url{https://steve-zeyu-zhang.github.io/DiabetesDiagnosis}}}

\end{abstract}
\section{Introduction}
\label{sec:intro}

Diabetes Mellitus (DM) is a chronic disease, originating from the Greek word \textit{diabetes}, characterized by persistently high blood glucose levels \cite{DM}. It adversely affects the heart, blood vessels, eyes, kidneys, and nerves, doubling the risk of vascular disorders in individuals with diabetes \cite{risk}. Evidence suggests a strong association between diabetes and certain malignancies (e.g., liver cancer) and other non-vascular illnesses \cite{2,3,4}. By the end of 2019, diabetes became the ninth leading cause of death, rising by 70\% since 2000, with an 80\% increase in male fatalities \cite{5}. Diabetes directly caused 1.5 million deaths worldwide, 48\% before the age of 70 \cite{6}. Currently, 37.3 million people in the US, or 11.3\% of the population, have diabetes, with 8.5 million undiagnosed individuals \cite{7}. Early diagnosis and treatment are crucial to prevent health risks as a "Silent Killer" \cite{8,9}. Implementing accurate prediction and monitoring approaches can significantly reduce the risk of developing the disease \cite{11}.

Currently, the majority of methods for predicting and diagnosing diabetes still rely on blood glucose level measurement \cite{49}. Specifically, invasive blood glucose laboratory tests and glucometers are standard solutions for glucose monitoring at hospitals and homes, respectively \cite{12}. Although these methods can provide relatively accurate test results, some evident disadvantages, such as stringent demand for skills and types of equipment, prohibitive costs, time-consuming, and the pain associated with testing, cannot be ignored \cite{13}.

In comparison, machine learning and deep learning-based diabetes diagnosis gather data from real-world datasets, which does not require special instruments and has the advantages of low cost and high efficiency. The most commonly used dataset is the Pima Diabetes dataset carried out by the US National Institute of Diabetes and Digestive and Kidney Diseases (NIDDK) \cite{16,17,18} and available at the University of California Irvine Machine Learning Repository \cite{19}. There are other well-known datasets that can be used in Diabetes diagnosis, such as CDC BRFSS 2015 Diabetes Health Indicators Dataset \cite{20,21,22}, and BIT Mesra Diabetes Dataset 2019 \cite{25,26}. In the datasets utilized, the types of data encompass various symptomatic observations and body measurements associated with diabetes. The outcome or target variable is binary, indicating whether an individual has diabetes (positive) or does not have diabetes (negative). It is important to note that all the features employed in the datasets are acquired through non-invasive methods. However, it is crucial to acknowledge the absence of the gold standard diagnostic criteria for diabetes, such as blood glucose level, in the data collection process.

Some machine learning methods have been proposed for diabetes prediction and achieved some encouraging progress with the three datasets mentioned earlier, for example, Classwise k Nearest Neighbor (CkNN) from Christobel's work \cite{32} and regression neural network (GRNN) from Kayaer's work \cite{20}. They tend to be statistical learning methods or feed-forward neural networks. Moreover, most of these models, such as Multi-Layer Feed Forward Neural Networks (MLFNN) from Kumar's work \cite{36}, have been only validated on a single dataset without indicating sensitivity and specificity, which leads to a relatively limited persuasiveness. Smith et al. \cite{27} designed a prediction model based on an early neural network model, ADAP \cite{28,29}, which is an adaptive learning method that generates and executes digital analogs of perceptron-like devices. They tested it on the Pima Indians diabetes dataset, and the performance was measured by sensitivity and specificity, which achieved 76\% at the crossover point. Wahba et al. \cite{37} applied two models on diabetes datasets, penalized log-likelihood smoothing spline analysis of variance (PSA) and Generalized Linear Models (GLIM) \cite{38}, which achieved accuracies of 72\% and 74\%, respectively. Breault et al. \cite{30} implemented a data mining algorithm, Rough sets \cite{31}, with the standard/tuned voting method (RSES) on the Pima diabetes dataset. Out of 392 complete cases, the model achieved a predictive accuracy of 73.8\% with a 95\% CI of (71.3\%, 76.3\%). Christobel et al. \cite{32} addressed the missing value in the Pima diabetes dataset using the mean method and implemented a new Classwise k Nearest Neighbor (CkNN) algorithm for the prediction of diabetes. Through 10-fold cross-validation, the algorithm has achieved an accuracy of 78.16\%. Kumari et al. \cite{33} proposed a classifier using Support Vector Machine (SVM) on Pima Indians diabetes dataset. The experimental results obtained an accuracy of 75.5\% for RBF kernel SVM and 78.2\% for SVM classification. Ahmad et al. \cite{34} designed a hybrid method that consists of an improved genetic algorithm (GA) for simultaneous parameter tuning and feature selection and a multi-layer perceptron (MLP) for classification. The model they developed obtained an accuracy of 80.4\% on the Pima dataset. Kayaer et al. \cite{35} developed a model based on General Regression Neural Network (GRNN), which consists of an input layer, two hidden layers (32 and 16 neurons, respectively), and an output layer with only one neuron. The classifier was examined on the Pima Indian dataset and achieved an accuracy of 80.21\%. Kumar et al. \cite{36} developed a classification model based on Multi-Layer Feed Forward Neural Networks (MLFNN), and achieved 81.73\% accuracy on the Pima diabetes dataset using the mean method for missing values. Polat et al. \cite{48} developed a classification model on the Pima dataset using Generalized Discriminant Analysis combined with Least Square Support Vector Machine (GDA-LS-SVM). Using 10-fold cross-validation, they achieved 79.16\% accuracy.

However, these methods may also have some imperfections. For instance, some methods, such as the ADAP algorithm from Smith's work \cite{28,29} and the Rough set algorithm from Breault's work \cite{30}, may have trained with imbalanced data directly without using the proper data preprocessing methods. This might lead the classifier to be biased toward the majority (negative) class and result in low sensitivity. Others may be designed a model that is not powerful enough or used an inappropriate model as the backbone for binary classification.

The primary objective of this paper is to propose and develop a deep-learning model and pipeline specifically designed for diabetes diagnosis. Our focus lies in leveraging data obtained through non-invasive methods as the sole input for our model. By solely relying on non-invasive data collection approaches, we aim to enhance the practicality and feasibility of the proposed solution for real-world applications. The developed model and pipeline strive to achieve accurate and reliable diabetes diagnosis based solely on non-invasive data, thereby mitigating the need for invasive diagnostic procedures and improving patient experience and convenience. We proposed a model based on Back Propagation Neural Network (BPNN) combined with batch normalization. The main contribution of this paper could be summarized as follows.

\begin{itemize}

\item We improved the sensitivity through implementing undersample-balancing in the procedure of data preprocessing.

\item We proposed a deep learning model based on Back Propagation  Neural Network (BPNN) for diabetes diagnosis. Specifically, by updating losses and biases through backward propagation, the accuracy of samples that are difficult to classify in some datasets has also been improved substantially.

\item We conduct experiments on four distinct real-world datasets with different features and dimensions. The horizontal comparison of the results indicates the superior performance of BPNN in terms of accuracy, among other approaches.

\end{itemize}

\section{Methodology}
\label{sec:method}

\begin{figure}[h]
    \centering
    \includegraphics[width=1\linewidth]{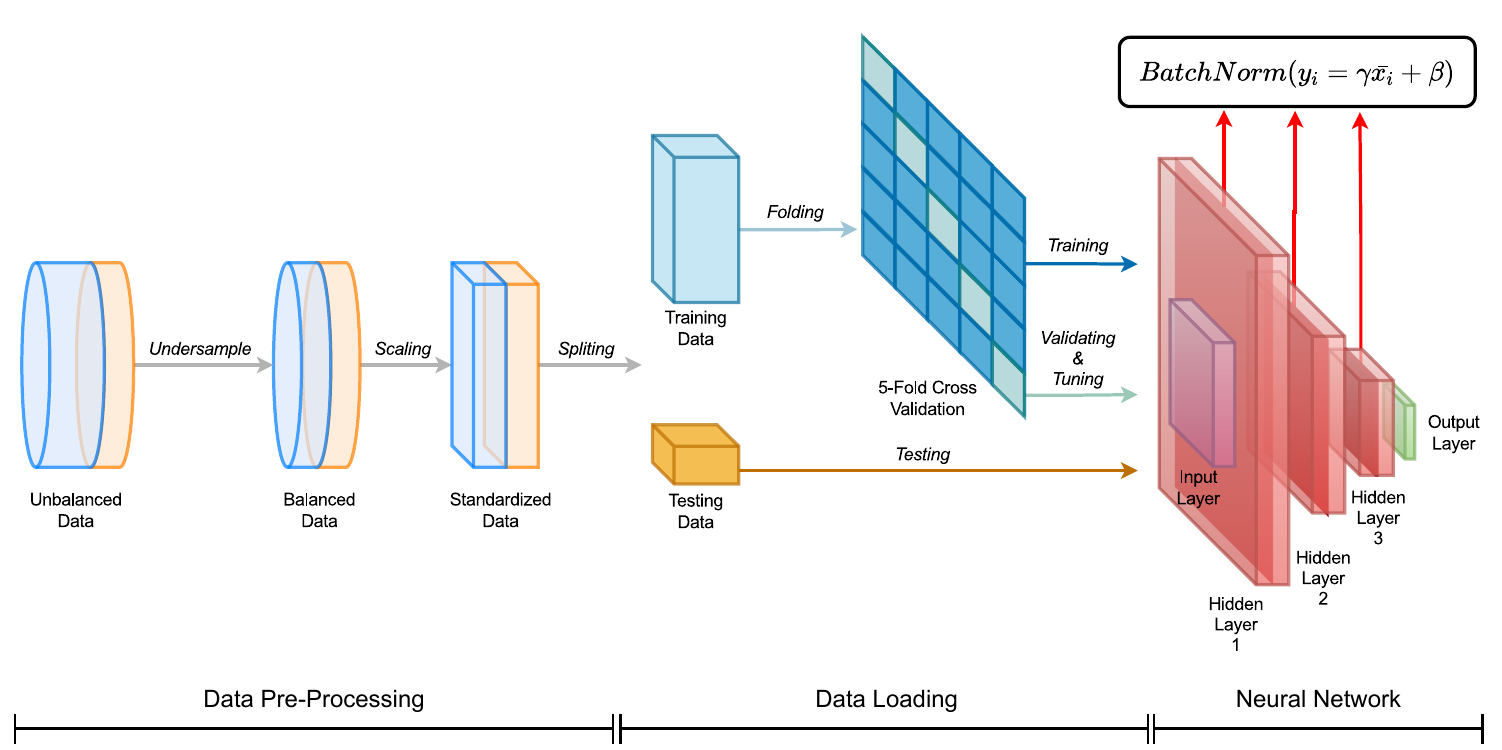} 
    \caption{Workflow of Proposed Method: The pipeline encompasses crucial components, including data undersampling to address class imbalance in the dataset. The Workflow of our proposed method illustrates the data scaling procedure for effective feature normalization. The backbone of the pipeline consists of a Back Propagation Neural Network (BPNN) architecture, enhanced with batch normalization, to facilitate automatic diabetes diagnosis. This comprehensive pipeline demonstrates potential for accurate and automated diabetes classification.}
    \label{fig:3}
\end{figure}

\begin{figure}[h]
    \centering
    \includegraphics[width=0.7\linewidth]{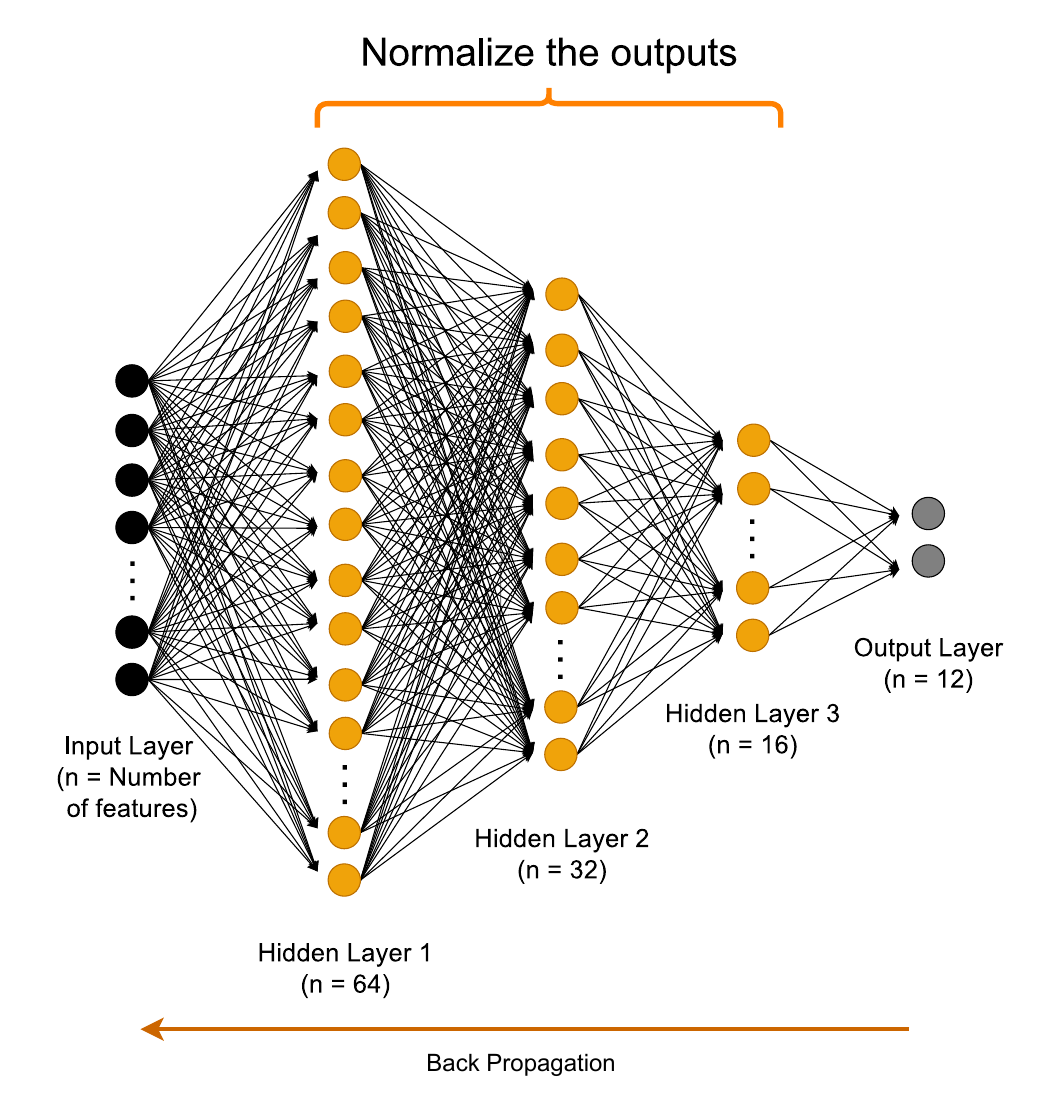} 
    \caption{Structure of BPNN}
    \label{fig:2}
\end{figure}

In this section, we outline our deep learning model for diabetes diagnosis and its structure. We chose a Back-Propagation Neural Network (BPNN) due to its superior representation and feature extraction capabilities compared to other statistical machine learning methods.
The BP algorithm works by iteratively updating the network's weights and biases based on the error between the predicted outputs and the actual targets from a set of training examples. The algorithm starts with a forward pass (feedforward) to compute the activations of each neuron in the network and then calculates the output error. It then propagates this error backward through the layers (backpropagate error), computing the errors for each neuron in each layer. Finally, the gradients of the cost function with respect to the weights and biases are computed using the errors, which are used to update the weights and biases in the network, thereby improving its ability to diagnose diabetes accurately. 
The BPNN is shown in Figure \ref{fig:2}, which is built up by full connections of an input layer, three hidden layers, and an output layer. The input layer possesses the same number of neurons as the features. The number of neurons in the hidden layers is 64, 32, and 16, respectively. Eventually, we have two neurons in the output layer referring to the two output classes.

In Figure \ref{fig:2}, each fully connected arrow stands for a feed-forward process. The sigmoid activation function has been implemented for each layer, and the output of each layer has been normalized.
We implement batch normalization \cite{42} to improve the training speed and stability, as well as to mitigate the issue of internal covariate shift, thus enhancing the overall performance of our BPNN for diabetes diagnosis.
The algorithm computes the mean and variance of inputs within a batch of training samples and then normalizes the inputs by subtracting the mean and dividing by the square root of the variance. This normalization step helps stabilize and speed up the training process. During inference, batch normalization uses running averages of the batch mean and variance to normalize the inputs, along with scaling and shifting parameters to obtain the final outputs of the layers, ensuring the network performs well on new, unseen data.

For the back propagation, the figure demonstrates a standard or typical MLP, not a Single-Layer perceptron. For standard MLP, it uses BP to update weight \& bias. We used the cross-entropy loss as the loss function and the adaptive moment estimation (ADAM) to search for the minima of the loss function.

The hyperparameter mentioned above is determined by grid search, which is explained in detail in Section \ref{sec:tuning}. In this process, a predefined set of hyperparameter values is defined for each hyperparameter (e.g., hidden layers, activation function, optimizer, mini-batch size), and the model's performance is evaluated for all possible combinations of these values using cross-validation. The combination that results in the best performance metric (e.g., accuracy, loss) on the validation set is then selected as the optimal set of hyperparameters for the model. Utilizing grid search, we employed an exhaustive search technique to identify the optimal hyperparameter configuration for the proposed Back Propagation Neural Network (BPNN) model. This systematic approach enabled us to maximize the performance of the BPNN by selecting the combination of hyperparameters that yielded the highest performance metrics. 

\section{Experiment}
\label{sec:experiment}

In this section, we evaluate the effectiveness of BPNN model on Pima Indian diabetes dataset and compare it with some statistical learning methods, other deep learning methods, and some existing methods done by related works. 

The stages of the experiment could be generally described as (1) Data Preprocessing, (2) Hyperparameter tuning of BPNN, and (3) Validation, which is shown in Figure \ref{fig:3}. The proposed pipeline's workflow involves three main steps for improving the performance of the model in handling unbalanced data. Firstly, an undersampling technique is applied to balance the class distribution in the dataset. Secondly, standardization is performed to scale the data, ensuring consistency in feature magnitudes. Lastly, the processed data is used to train a Back Propagation Neural Network (BPNN) model, adopting a five-fold cross-validation approach to assess its performance and ensure robustness in the evaluation process.

\subsection{Data Preprocessing}

\subsubsection{Overview of Dataset}

\begin{figure}[h]
    \centering
    \includegraphics[width=0.7\linewidth]{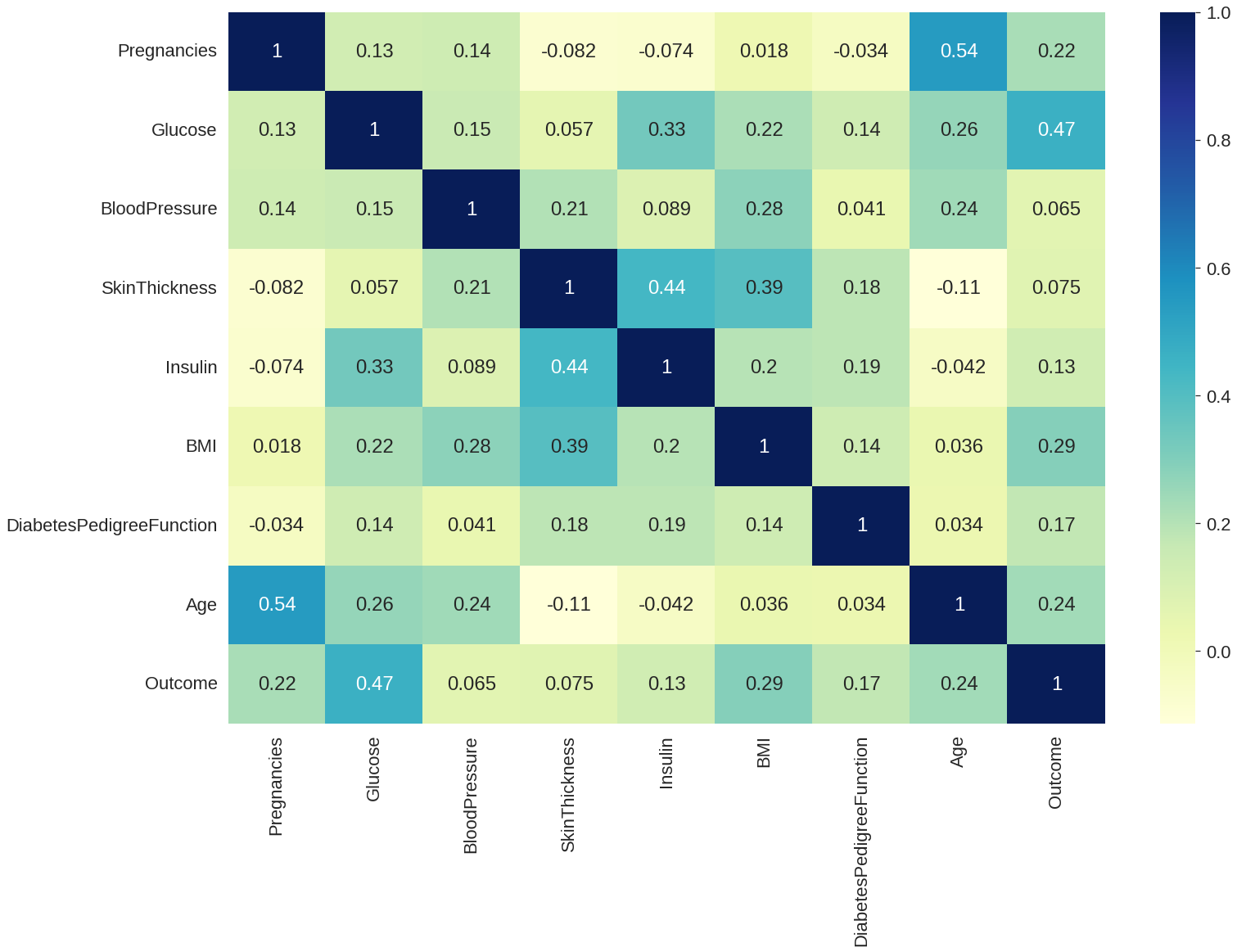} 
    \caption{Correlation matrix of Pima dataset: The figure displays the correlation matrix of the Pima dataset, providing a visual representation of the interrelationships between the variables within the dataset.}
    \label{fig:4}
\end{figure}

Pima Indian diabetes dataset is provided by National Institute of Diabetes and Digestive and Kidney Diseases (NIDDK) and the Applied Physics Laboratory of the Johns Hopkins University \cite{16}. The dataset provided 768 females at least 21 years old of Pima Indian heritage who responded to the survey. The dataset consists of several medical predictors (i.e. independent variables) and a target (dependent) variable, Outcome. Independent variables include the number of pregnancies the sample has had, their age, BMI, blood pressure (BP), insulin level, and so on. The correlation matrix of Pima dataset is shown in Figure \ref{fig:4}. Based on specific diagnostic metrics present in the dataset, the goal of the dataset is to diagnostically forecast whether a patient has diabetes or not.

\subsubsection{Data Balancing}

When one class dominates the other classes in a dataset relative to the target class variable, the dataset is said to be imbalanced \cite{39}. However, classification algorithms are designed to assume that the dataset is balanced \cite{39}. When a classifier is trained using an imbalanced dataset, it will probably be biased towards the majority class, which means that the performance of the classifier will be better at predicting the majority class than the minority class \cite{39}. Eventually, it will result in low sensitivity. Thus, an imbalanced dataset will introduce bias during training. Therefore, balancing imbalanced datasets is one of the most essential methods in data preprocessing since it will help reduce bias in the prediction, and thereby enhance the performance of the classifier.

The initial Pima dataset exhibited an imbalanced distribution, comprising 268 positive instances (with diabetes) and 500 negative instances (without diabetes). The Pima Indian diabetes dataset is a highly imbalanced data since the size of the negative class is significantly larger than the size of the positive class. To address this class imbalance, we applied a data undersampling technique. Undersampling is a technique that balances the dataset by randomly reducing the size of the majority class until reaching the size of the minority class. Despite it might discard some samples from the original dataset, it will not introduce any bias to training and is considered to be one of the most widely used data balancing methods. Consequently, the dataset was rebalanced, resulting in an equal number of instances, namely 268 instances in each class.

\subsubsection{Data Scaling}

It is well known that most machine learning methods evaluate the data distance or similarity (e.g., Euclidean distance) to make inferences and predictions. However, few features are measured on the same scale. Specifically, the majority of the features are either different in magnitudes or different in units. Hence, scaling the data will bring every feature the same contribution to the classification, which will enhance the performance of classification algorithms \cite{40}. Scaling will also reduce the time spent training. If the values of the features are closer to each other, it will accelerate the process for the classifier to understand the data and speed up the process of convergence of gradient descent \cite{41,42}.

\begin{figure}[h]
    \centering
    \includegraphics[width=0.8\linewidth]{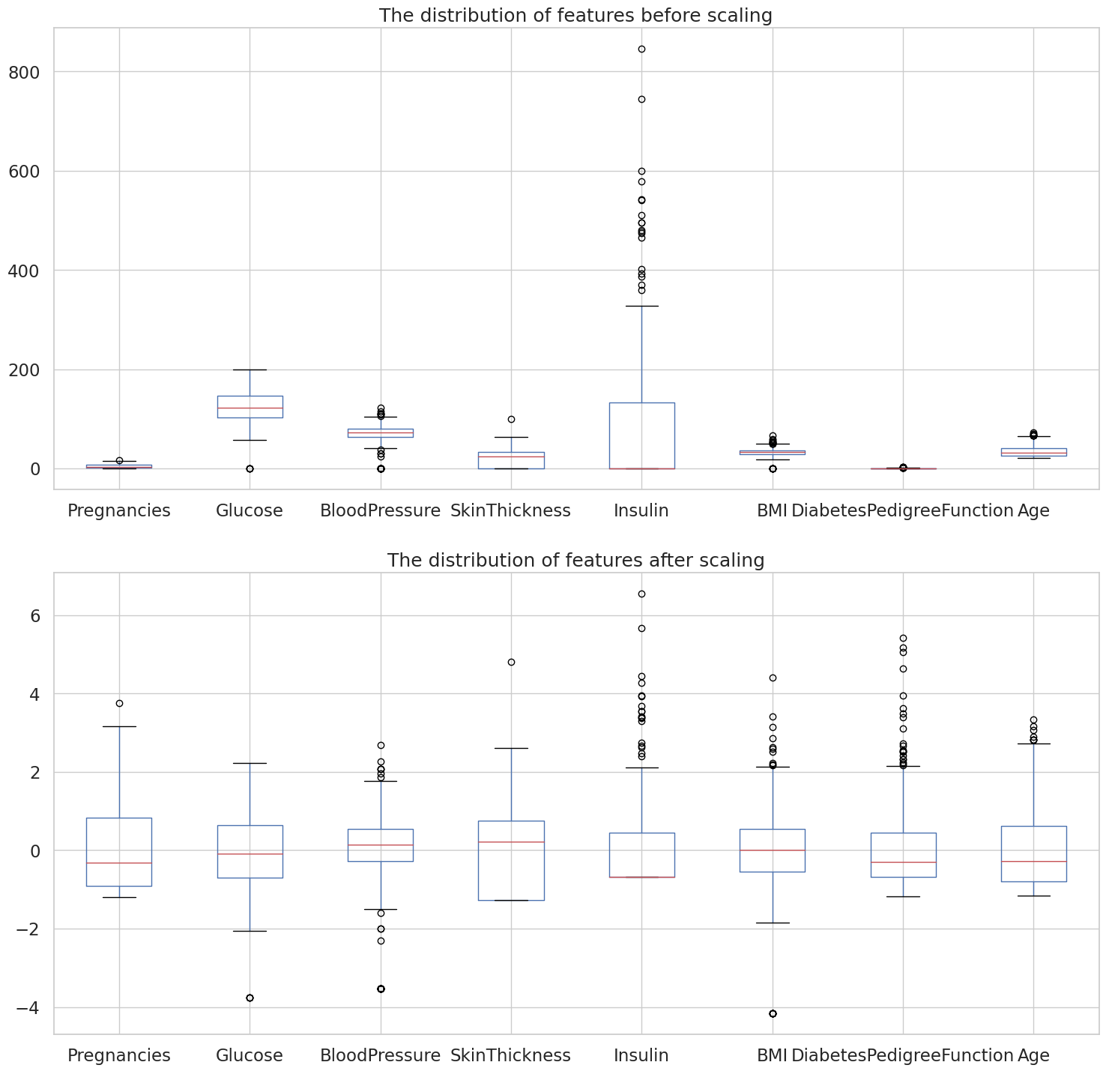} 
    \caption{The figure displays the feature distributions for diabetes diagnosis in the dataset before (top sub-figure) and after (bottom sub-figure) scaling using standardization. Standardization has successfully transformed the features to a comparable magnitude, resulting in a more uniform distribution, facilitating the training process and enhancing the performance of the Back Propagated diabetes diagnosis model.}
    \label{fig:6}
\end{figure}

There are two major approaches for scaling the data: normalization and standardization. Hence, we choose standardization as our scaling method since it does not harm the position of outliers, wherein the normalization captures all the data in a certain range. The distribution of the features before and after scaling is shown in Figure \ref{fig:6}.

For standardization, we have
\begin{equation}
  x_{stand} = \frac{x-\Bar{x}}{\sigma(x)}
  \label{eq:19}
\end{equation}
\noindent where $\sigma(x)$ refers to the standard deviation of $x$.

\subsubsection{Data Visualization}

\begin{figure}
    \centering
    \includegraphics[width=0.9\linewidth]{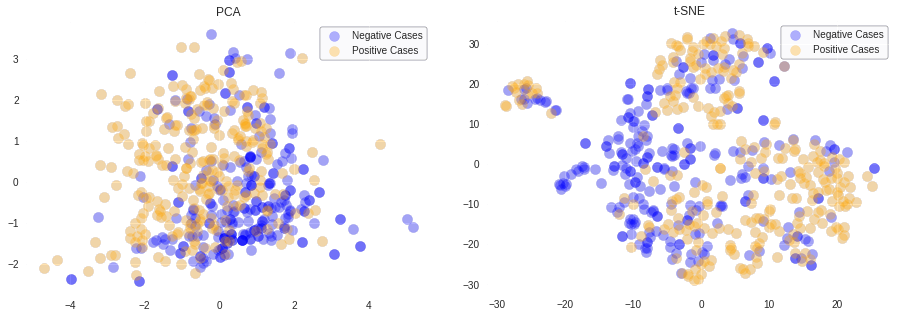} 
    \caption{The plot compares the distribution of positive and negative samples using two methods, PCA (linear dimensionality reduction) and t-SNE (nonlinear dimensionality reduction), providing a comprehensive visualization of their distribution in the dataset.}
    \label{fig:7}
\end{figure}

The visualization process involved two dimensionality reduction techniques: Principal Component Analysis (PCA) and t-Distributed Stochastic Neighbor Embedding (t-SNE). PCA captured the most important information in the data, presenting it in a way that highlights similarities and differences. t-SNE, on the other hand, focused on visualizing high-dimensional data by creating a probability distribution that emphasized similarities and minimized the divergence between high and low-dimensional representations. While t-SNE is better suited for non-linear data, it comes with a higher computational complexity. Both PCA and t-SNE were employed to reduce the data into two dimensions for visualization purposes, which is shown in Figure \ref{fig:7}.

\subsection{Hyperparameter Tuning}
\label{sec:tuning}

Grid search, also known as parameter sweep, is a hyperparameter optimization method performed by a thorough search across a manually chosen subset of a learning algorithm's hyperparameter space. An evaluation on a hold-out validation set or cross-validation on the training set are two common ways to measure performance metrics for grid search algorithms. Prior to conducting a grid search, manually established boundaries and discretization may be required since the parameter space of a classifier may comprise real-valued or unbounded value spaces for some parameters.

To implement grid search for hyperparameter tuning, we need to determine a subset of hyperparameter space as the grid search dictionary. Eventually, we derived the optimal hyperparameters for BPNN, which is demonstrated in Figure \ref{table:2}.

\begin{table}[h]
\centering %
\caption{Chosen subset of hyperparameter space, and optimal hyperparameters for BPNN, which are labeled in red.} %
\label{table:2} %
\begin{tabular}{c c} %
\hline\hline %
Hyperparameter & Value  \\ %
\hline %
Hidden Layer & [16,8,4],[32,16,8],\textcolor{red}{[64,32,16]} \\ %
Activation & \textcolor{red}{Sigmoid}, ReLU \\
Optimizer & SGD, \textcolor{red}{Adam} \\
Mini Batch & 8, \textcolor{red}{16}, 32 \\

\hline %
\end{tabular}
\end{table}

\subsection{Validation}

\subsubsection{Cross Validation}

K-fold cross-validation is one of the most widely used approaches for parameter tuning during training and performance evaluation of a classifier. 
The dataset has been first split into two subsets, 80\% for training and 20\% for testing purposes. During the training process, the training data is randomly split into 5 folds, 
For each iteration, we use four of them for training and one for validating, and these training data were used to train and tune the hyperparameters of the BP neural network. Once the iteration is completed, the model has already been fine-tuned and will be validated using the testing data.

\subsubsection{Evaluation Metrics}

Several evaluation metrics were used to test the performance of the developed model. One of the most well-known indicators is accuracy, which is defined as the percentage of all identifications that are actually correct. Moreover, to ensure the model is not biased towards a single class, we also use sensitivity and specificity, which are the true positive rate and true negative rate, respectively.

\subsection{Results}

We investigate the performance of the proposed BPNN on testing data and achieved 89.81\% for accuracy, 89.29\% for sensitivity, and 90.38\% for specificity. 
We also compared our BPNN model with several machine learning methods. Moreover, we also compared with some of the best performing related works, including the Classwise k Nearest Neighbor (CkNN) from Christobel's work \cite{32}, the improved genetic algorithm and multi-layer perceptron (GA-MLP) from Ahmad's work \cite{34}, General
regression neural network (GRNN) from Kayaer's work \cite{20}, Multi-Layer Feed Forward Neural Networks (MLFNN) from Kumar's work \cite{36}, and Generalized Discriminant Analysis combined Least Square Support Vector Machine (GDA-LS-SVM) from Polat's work \cite{35}. 
The results on the Pima Indian diabetes dataset and other datasets are shown in Table \ref{table:4}. Our proposed BPNN outperformed CkNN by 11.65\%, GDA-LS-SVM by 10.65\%, GA-MLP by 9.41\%, GRNN by 9.6\%, and MLFNN by 8.03\% in the Pima diabetes dataset. The underperformance of the least performing models compared with our model can be attributed to two main factors. Firstly, these models did not utilize data balancing and scaling techniques, resulting in an unbalanced training data that tends to favor the major class, thereby significantly impacting their performance, as seen in GA-MLP, GRNN, and MLFNN. Secondly, traditional statistical machine learning methods, such as CkNN and GDA-LS-SVM, lack the capability to extract deep abstract features, which hinders their performance when compared to deep neural networks. Consequently, the deep neural network serves as an effective encoder for feature extraction, contributing to the classifier's superior performance. It is obvious that the proposed method has significantly improved the accuracy of diabetes diagnosis compared with other machine learning methods.

\begin{table*}[h]
\centering
\caption{Comparative results on different datasets with various models. The cells with `-' indicate that certain comparative studies did not assess their models on specific datasets.} %
\label{table:4} %
\resizebox{1\textwidth}{!}{
\begin{tabular}{c|ccc|ccc|ccc}
\hline
\multicolumn{1}{c}{} & \multicolumn{3}{|c}{NIDDK Pima Indian Diabetes Dataset} & \multicolumn{3}{|c}{CDC BRFSS2015 Database} & \multicolumn{3}{|c}{BIT Mesra Diabetes Dataset}  \\ \cline{1-10}
 Method & Test. Acc. & Sensitivity & Specificity & Test. Acc. & Sensitivity & Specificity & Test. Acc. & Sensitivity & Specificity \\ \hline
LDA &  0.7222 & 0.6721 & 0.7872  & 0.7416 & 0.7767 & 0.7064 &  0.8868 	&0.9048 	&0.875  \\
KNN &  0.8148 & 0.7705 & 0.8723  &  0.7376 & 0.7952 & 0.6792  &  0.9151 	&0.9524 	&0.8906  \\
Logistic Regression &  0.6852&	0.6230&	0.7660  &  0.7418 	&0.7685 	&0.7148  &  0.8396 	&0.9286 	&0.7813  \\
SVM &  0.7130&	0.6393	&0.8085  &  0.7411 	&0.7906 	&0.6908 &   	0.8491 	&0.8571 	&0.8438  \\
Decision Trees &  0.7037&	0.6885&	0.7234  &  0.7364 	&0.7622 	&0.7102  &  0.9057 	&0.9524 	&0.875  \\
Random Forest &  0.7222&	0.7213&	0.7234  &  0.7304 	&0.7673 	&0.6930  &  0.8774 &	0.9286 &	0.8438  \\
Bagging &  0.6944&	0.6230&	0.7872  &  0.7477 	&0.7983 	&0.6964  &  0.8868 	&0.9524 	&0.8438  \\
XGBoost &  0.7870&	0.7377&	0.8511  &  0.7505 	&\textcolor{red}{0.7987} 	&0.7017  &  0.9245 & 	0.9524& 	0.9062  \\
K-Means Clustering &  0.6481&	0.4590&	0.8936  &  0.6653 	&0.5069 	&\textcolor{red}{0.8259}  &  0.7264 	&0.4762 	&0.8906  \\
SOM &  0.7130 &	0.6721 &	0.7660  &  0.6611 	&0.5118 	&0.8125  & 0.6698 	&0.5714 	&0.7344  \\
ResNet-14 &  0.7963&	0.7049&	\textcolor{red}{0.9149}  &  0.7492& 	0.7790& 	0.7187  &  0.9245 	&0.9524 	&0.9063 \\
ResNet-50 &  0.7870 &	0.7706 &	0.8085  &  0.7442& 	0.7722 &	0.7158  &  0.9151 	&0.9286 	&0.9062  \\
CkNN &  0.7816 & 0.6184 & 0.8738  & --- & --- & --- &  --- & --- & ---  \\
GDA-LS-SVM 3 &  0.7916 & 0.8333 & 0.8205 & --- & --- & --- &  --- & --- & ---  \\
GA-MLP &  0.8040 & --- & ---  & --- & --- & --- &  --- & --- & ---  \\
GRNN &  0.8021 & --- & ---  & --- & --- & --- &  --- & --- & --- \\
MLFNN &  0.8173 & --- & ---  & --- & --- & --- &  --- & --- & --- \\
\textbf{DiabetesNet (Ours)} &  \textcolor{red}{\textbf{0.8981}} & \textcolor{red}{\textbf{0.8929}} & \textbf{0.9038}  &  \textcolor{red}{\textbf{0.7549}} 	&\textbf{0.7977}	&\textbf{0.7112} &  \textcolor{red}{\textbf{0.9528}} &	\textcolor{red}{\textbf{1.0}} 	& \textcolor{red}{\textbf{0.9219}} \\
\hline
\end{tabular}
}
\end{table*}

\section{Conclusion}

In this study, we introduced an innovative diabetes diagnosis model that leverages the Back Propagation Neural Network (BPNN) in synergy with batch normalization. Our model presents a noteworthy advancement in enhancing the accuracy of diabetes diagnosis across authentic datasets. The substantial performance improvement demonstrated not only surpasses related models but also potentially positions it as a benchmark, signifying its pivotal role in shaping the landscape of diabetes diagnosis. Despite limited dataset size and features, our method showed promising results across multiple datasets for diabetes diagnosis. Moving forward, our future work will involve refining and validating our approach with more comprehensive datasets to enhance its robustness and generalizability. Additionally, we aim to improve our diagnostic approach through data processing refinement and feature engineering.


\begin{thebibliography}{10}
\providecommand{\url}[1]{\texttt{#1}}
\providecommand{\urlprefix}{URL }
\providecommand{\doi}[1]{https://doi.org/#1}

\bibitem{34}
Ahmad, F., Isa, N.A.M., Hussain, Z., Osman, M.K.: Intelligent medical disease diagnosis using improved hybrid genetic algorithm - multilayer perceptron network. Journal of Medical Systems  \textbf{37}, ~9934 (2013)

\bibitem{40}
Ahsan, M.M., Mahmud, M.A.P., Saha, P.K., Gupta, K.D., Siddique, Z.: Effect of data scaling methods on machine learning algorithms and model performance. Technologies  \textbf{9}(3), ~52 (2021)

\bibitem{16}
Bennett, P.H., Burch, T.A., Miller, M.: Diabetes mellitus in american (pima) indians. Lancet  \textbf{298}(7716),  125--128 (1971)

\bibitem{30}
Breault, J.L.: Data mining diabetic databases: Are rough sets a useful addition? The 33rd Symposium on the Interface, Computing Science and Statistics (2001)

\bibitem{19}
of~California Irvine Machine Learning~Repository, U.: Pima indians diabetes dataset

\bibitem{32}
Christobel, Y.A., Sivaprakasam, P.: A new classwise k nearest neighbor (cknn) method for the classification of diabetes datase. International Journal of Engineering and Advanced Technology  \textbf{2}(3),  396–400 (2013)

\bibitem{8}
Clarke, S., Foster, J.: A history of blood glucose meters and their role in self-monitoring of diabetes mellitus. British Journal of Biomedical Science  \textbf{69}(2),  83--93 (2012)

\bibitem{risk}
Collaboration, T.E.R.F.: Diabetes mellitus, fasting blood glucose concentration, and risk of vascular disease: a collaborative meta-analysis of 102 prospective studies. Lancet  \textbf{375}(9733),  2215--2222 (2010)

\bibitem{6}
of~Disease Collaborative~Network, G.B.: Global Burden of Disease Study 2019 Results. Institute for Health Metrics and Evaluation (2020)

\bibitem{20}
for Disease~Control, C., (CDC), P.: 2015 behavioral risk factor surveillance system (brfss) survey data and documentation

\bibitem{7}
for Disease~Control, C., Prevention: National Diabetes Statistics Report (2022)

\bibitem{2}
El-Serag, H., Hampel, H., Javadi, F.: The association between diabetes and hepatocellular carcinoma: A systematic review of epidemiologic evidence. Clinical Gastroenterology and Hepatology  \textbf{4}(3),  369--380 (2006)

\bibitem{9}
Galaviz, K., Narayan, K., Lobelo, F., Weber, M.: Lifestyle and the prevention of type 2 diabetes: A status report. American Journal of Lifestyle Medicine  \textbf{12}(1),  4–20 (2018)

\bibitem{3}
Huxley, R., Ansary-Moghaddam, A., Berrington~de Gonzalez, A., Barzi, F., Woodward, M.: Type-ii diabetes and pancreatic cancer: a meta-analysis of 36 studies. British Journal of Cancer  \textbf{92},  2076–2083 (2005)

\bibitem{42}
Ioffe, S., Szegedy, C.: Batch normalization: accelerating deep network training by reducing internal covariate shift. International Conference on Machine Learning (ICML) (2015)

\bibitem{39}
Jadhav, A., Mostafa, S.M., Elmannai, H., Karim, F.K.: An empirical assessment of performance of data balancing techniques in classification task. Applied Sciences  \textbf{12}(8), ~3928 (2022)

\bibitem{4}
Kasper, J.S., Giovannucci, E.: A meta-analysis of diabetes mellitus and the risk of prostate cancer. Cancer Epidemiology, Biomarkers \& Prevention  \textbf{15}(11),  2056–2062 (2006)

\bibitem{35}
Kayaer, K., Yildirim, T.: Medical diagnosis on pima indian diabetesusing general regression neural networks. Proceedings of the international conference on artificial neural networks and neural information processing  \textbf{26-29},  181--184 (2003)

\bibitem{17}
Knowler, W.C., Bennett, P.H., Hamman, R.F., Miller, M.: Diabetes incidence and prevalence in pima indians: a 19-fold greater incidence than in rochester, minnesota. American Journal of Epidemiology  \textbf{108}(6),  497--505 (1978)

\bibitem{18}
Knowler, W.C., Pettitt, D.J., Savage, P.J., Bennett, P.H.: Diabetes incidence in pima indians: contributions of obesity and parental diabetes. American Journal of Epidemiology  \textbf{113}(2),  144--156 (1981)

\bibitem{36}
Kumar, S., Bhusan, B., Singh, D., Choubey, D.K.: Classiﬁcation of diabetes using deep learning. International Conference on Communication and Signal Processing (2020)

\bibitem{33}
Kumari, V.A., Chitra, R.: Classification of diabetes disease using support vector machine. International Journal of Engineering Research and Applications  \textbf{3}(2),  1797--1801 (2013)

\bibitem{12}
Mathew, T.K., Tadi, P.: Blood Glucose Monitoring. Treasure Island (FL): StatPearls Publishing (2021)

\bibitem{38}
McCullagh, P., Nelder, J.A.: Generalized Linear Models. CRC Press (1983)

\bibitem{5}
Organization, W.H.: Global Health Estimates 2019: Deaths by Cause, Age, Sex, by Country and by Region  (2020)

\bibitem{31}
Pawlak, Z.: Rough sets. International Journal of Computer \& Information Sciences  \textbf{11},  341–356 (1982)

\bibitem{48}
Polat, K., Günes, S., Arslan, A.: A cascade learning system for classification of diabetes disease: Generalized discriminant analysis and least square support vector machine. Expert Systems with Applications  \textbf{34}(1),  482--487 (2008)

\bibitem{DM}
Sapra, A., Bhandari, P.: Diabetes Mellitus. Treasure Island (FL): StatPearls Publishing (2022)

\bibitem{11}
Shi, B.: The importance and strategy of diabetes prevention. Chronic Diseases and Translational Medicine  \textbf{2}(4),  204–207 (2016)

\bibitem{41}
Singh, D., Singh, B.: Investigating the impact of data normalization on classification performance. Applied Soft Computing  \textbf{97},  105524 (2020)

\bibitem{29}
Smith, J.W.: Adap ii, an adaptive routine for the larc computer. Navy Management Office  (1962)

\bibitem{28}
Smith, J.W.: Adap, an adaptive algorithm. Logistics Management Institute Technical Report  \textbf{1}, ~705 (1986)

\bibitem{27}
Smith, J.W., Everhart, J.E., Dickson, W.C., Knowler, W.: Using the adap learning algorithm to forecast the onset of diabetes mellitus. Proceedings of the Annual Symposium on Computer Applications in Medical Care p. 261–265 (1988)

\bibitem{22}
Teboul, A.: Diabetes health indicators dataset

\bibitem{25}
Tigga, N.P.: Diabetes dataset 2019

\bibitem{26}
Tigga, N.P., Garg, S.: Prediction of type 2 diabetes using machine learning classification methods. Procedia Computer Science  \textbf{167},  706--716 (2020)

\bibitem{13}
Villena~Gonzales, W., Mobashsher, A., Abbosh, A.: The progress of glucose monitoring—a review of invasive to minimally and non-invasive techniques, devices and sensors. Sensors  \textbf{19}(4), ~800 (2019)

\bibitem{37}
Wahba, G., Gu, C., Wang, Y., Chappell, R.: Soft classification, a.k.a. risk estimation, via penalized log likelihood and smoothing spline analysis of variance. The mathematics of generalization: the proceedings of the SFI/CNLS Workshop on Formal Approaches to Supervised Learning pp. 331--360 (1992)

\bibitem{21}
Xie, Z., Nikolayeva, O., Luo, J., Li, D.: Building risk prediction models for type 2 diabetes using machine learning techniques. Preventing Chronic Disease  \textbf{16} (2019)

\bibitem{49}
Yu, Z., Jiang, N., Kazarian, S.G., Tasoglu, S., Yetisen, A.K.: Optical sensors for continuous glucose monitoring. Progress in Biomedical Engineering  \textbf{3},  022004 (03 2021). \doi{10.1088/2516-1091/abe6f8}

\end{thebibliography}

\end{document}